\documentclass{jfm}
\usepackage{graphicx}
\usepackage{epstopdf, epsfig}

\shorttitle{Machine Learning for Arduino Controlled Pick ‘n’ Place Robotic Arm}
\shortauthor{Masoud Yousefi}

\title{Arduino Controlled Pick ‘n’ Place Robotic Arm}

\author{Masoud yousefi\aff{1}
  \corresp{\email{My35@uakron.edu}}}

\affiliation{\aff{1} University of Akron}

\begin{document}

\maketitle

\begin{abstract}
This document is an introduction to designing a multiple degree of freedom robotic manipulator.  The goal of the robot is to sort an object based upon the object’s color.  The robot will be a synthesis of several linkages, servo motors, an Arduino system, and an end-effector.  The design must be able to autonomously determine what object to move and where to place it. The robot will have a maximum reach of 24” horizontally. The operating conditions include being able to sort several different colored spheres within the area of a 180° up to 16” from the base of the arm. Sorting consists of picking up the object from a random position near the base and moving it to a storing location near the edge of the defined operating range. With the following performance criteria and a limited budget, the final design will enter a competition to determine the most robust, and effective robotic manipulator.  
\end{abstract}


\section{Introduction}
\subsection{Problem Statement}
The focus of this project is to develop a multiple degree of freedom robot arm that can detect an object based on color, and then place the object in a specific location.  The object’s color will define the final location of the object. The final design must be able to operate autonomously once the initial and final position of the object have been specified. The arm will consist of linkages, servo/stepper motors, computer vision system, and a microcontroller system.  A budget of 250 USD is the financial total the group has to work with when constructing the design.  Then, once a working model is developed the design will be subjected to Von Mises and Deformation analysis using ANSYS, as well predictive manufacturing costs analysis\citep{Batchelor59}.  
\subsection{Sponsor Background}
Professor Brown is experienced in computational dynamics, biomechanics, and a variety of aspects of manufacturing \citep{abtahi2020machine2,abtahi2020machine}. The goal of his project is to showcase the skills and methods utilized by design engineers in a variety of manufacturing and industrial positions. Designing a multiple degree of freedom robotic arm will require knowledge from many of the design courses in the mechanical engineering curriculum. Although the current curriculum introduces topics like control systems and kinematics, further research will be required to develop the project from the ground up. By the end of the project all group members should be able to demonstrate skills that are valuable to an engineer in a manufacturing or industrial setting.
\subsection{Literature Survey}
In order to alter the positioning of the robotic arm, there is a need for numerous DC motors \citep{bagheri2016thermohydrodynamic,acharya2019instability}. For this project application, a performance comparison between servo and stepper motors is vital to the overall performance and cost of the design.  The benefits and downfalls for each of the motors were investigated to make intelligent design decisions for the manipulator.
A stepper motor, given predetermined voltage pulses, moves in predictable and discrete angular increments (eg. 7.5 deg, 0.18 deg) \citep{azarhooshang2020review,abtahi2020turbulent}.  The small incremental steps of a stepper motor, as well as its ability to rotate a full 360 degrees give the design substantial accuracy. Stepper motors are often used in cheaper, open loop systems since there is no feedback generated from an encoder, but for this reason stepper motors run the risk of losing their position.  A loss in position typically occurs when the motor is loaded past its maximum torque rating.  Also, stepper motors have excellent holding torque (also called frictional torque) at zero and low RPMs. However, as the speed of the motor increases, the stepper motor’s torque output continues to decrease. Overall, step motors are a natural choice when design specs call for low-cost automation with accurate, open-loop control \citep{hooshyar2020energy,Lee71,Rogallo81}.
Servo motors on the other hand operate in closed-loop systems, where feedback is utilized by the controller. For a simple closed loop system, feedback is generated from an encoding device, and given back to the controller in order to make fine-tuned adjustments. Servo motors are often utilized since errors can be corrected by the controller [3]. Servo motors also require an amplifier which takes the signal from the controller and converts the signal to an amount of current that is delivered to the motor (Pulse Width Modulation).  Also, the output torque for a servo motor is relatively constant for low and medium speeds. However, as the servo motor approaches its maximum rated speed, the torque begins to drop significantly.  In summary, servo systems are typically costlier due to the encoding requirement but are best utilized when a wide range of speeds are needed. 
Also, in order for the robotic arm to locate an object, there is a need for a sensor apparatus. Based on the system boundaries of this project, an ultrasonic sensor will be used in order to locate the target object \citep{Ursell50,Ursell50,Miller91,abtahi2020integrated}. The benefits and downfalls of the ultrasonic sensor were also investigated when making the design decision.
The ultrasonic sensor is about 45x20x15mm in size with two piezoelectric transducers, one on each end and has a 4-pin connector. Two of the pins work to provide the sensor with 5 volts. A third pin is the trigger pin and the fourth pin, is used to read the results of the measurements \citep{Lee71,Worster92,abtahi2019porous}. The working current is 15mA. An ultrasonic sensor operates by emitting a conical ultrasonic wave (40,000 Hz) which travels through the air. If there is an object in front of its path, the wave will bounce back (echo) to the module. The waves travel time and speed (340 m/s) are then used to calculate the distance between the sensor and the object being detected. This concept can also be applied to measure the dimensions of an object \citep{Koch83,Linton92}.
The sensor is able detect and differentiate many types of materials such as wood, metal, plastic, and glass. Smooth and rigid materials are the easiest to detect because of their ability to reflect sound energy to the sensor. At maximum sensitivity, the ultrasound beam can reach an area of 38 inches across at 10 feet away from the sensor. One downfall of the ultrasonic sensor is that its waves can be affected by parameters such as humidity, ambient noise, and room temperature. 
\subsection{Design Criteria}
The design criteria for this project is based on the fundamentals of research and the implementation of said research. While there are currently numerous amounts of pick n’ place machines currently on the market, they are mostly specialized to perform a specific task. In order to meet the needs and goals of Professor Michael Brown, the robotic arm must be sourced and optimized in a cost and time efficient manner to accomplish the specific tasks at hand \citep{Martin80,Hwang70,esfahanian2014fluid}. 
The goal of this project is centralized around using a 6 Degree of Freedom (DoF) robotic arm. The arm must be able to pick and place a Jenga block. The arm must retrieve the block from a predetermined location, and then place it in a different predetermined location. These locations are subject to change, so the arm must be able to accommodate a wide range of position inputs. To limit the scope of the arm, the locations are confined within a 2” offset between a 12” to 14” radius. The arm must be able to sweep through at least 180 degrees of the 2” strip \citep{liu2020machine,brunton2020machine}. 
The arm’s effectiveness will be judged on four main criteria. These criteria include speed, accuracy, precision, and cost. Speed is defined as how quickly can the arm complete an entire cycle of picking up an object, placing item and then being prepared to repeat said tasks. Accuracy and precision are based on the certainty and repeatability of the arm’s final location. Ultimately, the design must be cost effective and meet a production run requirement of at least 500 devices per year at a profit of $30\%$ over the prototype cost. There is 250.00 USD budget for designing the arm provided by the sponsor.
A final area of consideration not listed directly within the judging criteria is safety. While the final design may not receive points for being safe to use; from an ethics standpoint, consideration must be taken into making the machine safe to use. This product will be available for individuals with no technical background. 
\section{Technical Content}
\subsection{Assumptions }

When a project requires designing a product that this flexible and adaptable to its environment, many aspects are left open ended. Assumptions are made to limit the potentially endless possibilities that that the final product must be able to overcome. For the design of a 6 DoF robotic arm, several key assumptions must be made.
It is assumed that the arm will be used on earth, so movement will be limited due the earth’s estimated 9.8 m/s2 gravitational constant. It will be assumed that the arm will be operated indoors around room temperature. Therefore, the arm will not be subject to harsh winds, extremely high/low temperature fluctuations, water submersion. 
It is assumed the arm will be used on a stationary, smooth, and non-inclined surface. Once the arm has begun moving, its range of motion will be completely free of obstruction besides the block and obstacles apart of the course. Finally, for functionality, it is assumed that the block will always be in the same initial orientation (i.e. standing, or lying flat) before it is picked up by the arm. The block and obstacles will both be inanimate objects.
\subsection{Metrics}
 
There will be five metrics used to evaluate the success of this project. The stability of the robotic arm over time and the arms ability to perform the designed task are two very important metrics to be measured. Building the project with a 250 budget, overall speed and accuracy, and aesthetic appeal to the public follow shortly after the aforementioned metrics. 
The first two metrics can be tested by grabbing an object and placing it onto a specified space on a plane. An ANSYS simulation will test the stability of the arm by assessing by the amount of stress it can endure. A cost analysis will be completed in order to evaluate the amount of money spent on building the robot arm. Successful trials will be timed and recorded in order to measure the speed of the arm. It should take the arm no longer than 10 seconds to pick and place the object. People who have seen the robotic arm will be asked to answer survey questions that will measure their interest in the robotic arm. A point based system can be used to determine the success of metrics. For Example:
\begin{table}
  \begin{center}
\def~{\hphantom{0}}
  \begin{tabular}{lccc}
      Criteria  & Rating  \etal \\[3pt]
      Ability to perform assigned task	& /10 \\
      System Stability	& /10 \\
      Speed and Accuracy	& /10 \\
      Meeting Budget	& /10 \\
      Aesthetic Appeal	& /10 \\
  \end{tabular}
  \caption{Example Table for Rating Metrics.}
  \label{tab:kd}
  \end{center}
\end{table}

\subsection{Proposed Solutions}
 
As mentioned earlier in this paper, the whole process of picking and placing the object should be autonomous and limited human interaction is allowed while robotic arm is performing the task. The main challenge in this project is to automate object detection process so it will be the primary focus of this section.  One possible solution is to implement vision system and pattern recognition techniques. Integrated robotic arms with vision systems are widely utilized in industrial sectors. In many applications, the arm is required to detect objects with different dimensions and shapes placed at random positions over working area. Besides, objects might not be positioned well separated and aligned since doing so is a waste of time and space3. Without a doubt, programming such a sophisticated system would require long hours of work and rigorous research. However, dimensions and appearance of the desired object, which is a Jenga block, is known. Moreover, we know that no other object with similar sizes and shape will be placed over working area. This information would significantly lessen complexity of machine vision algorithms and allow to use simple photocells instead of expensive high-resolution cameras.
One way to detect objects is through PixyCam. It is an image sensor board that uses an external camera to get raw image data, thus acting as a visual sensor. Many image sensors output a lot of data that can easily override processors, however Pixy comes with its own processor that filters data so that only the most relevant data is sent to the microcontroller. It works primarily with Arduino and has all Arduino libraries. Pixy is small enough to be mounted onto a robot arm if needed. It uses color signatures to detect objects (up to 7 color signatures). Another feature is that it can be taught to recognize specific objects based on its color. The PixyCam has open source software, making it easier to work with. 
As already described in the Design Criteria section, the ultrasonic sensor is a proximity sensor will be able to detect objects. An alternative is the infrared proximity sensor which works in a similar way. All objects emit heat energy i radiation. These objects give off radiation that radiates at the infrared wavelength, which can then be detected by an infrared sensor. This sensor does not generate or radiate energy in order to detect objects, making it a passive. It functions by sensing infrared radiation emitted by or reflected from objects. Both the ultrasonic and infrared sensors are considerably cheaper than the PixyCam. However, they can only detect objects, not differentiate them by their color. This sorting ability would have to be handled by a different component.  
Figure 1 represents the general steps in pick and place operation. The user selects dimensions and shape of the object as well as coordinates of final position through user interface. Next, vision system captures images of working area and extracts features of existing objects and matches with input data using Scale-Invariant Feature Transform (SIFT) algorithm3. In object localization phase, the system tries to find coordinates of the selected object by mapping 2D coordinates of images to real-world 2D coordiantes3. Based on the vision system’s output data, the micro controller manipulates the arm to object’s location.  Finally, robotic arm picks up the object and places it in designated position.   
\begin{figure}
  \centerline{\includegraphics{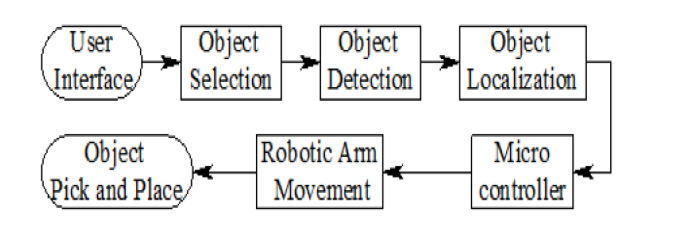}}% Images in 100% size
  \caption{Diagram of Pick and Place System.}
\label{fig1}
\end{figure}

One way to select the desired object and set final position is by an implementation of hand gesture method. In this method, two cameras are attached to top of the upper arm. One captures real-time images of working area and the other one is used to track movements of user’s finger. OpenCV is an open library source for computer vision that can be used to determine size, shape, color, distance to an object, etc. Contours of objects and user’s finger are created in OpenCV. The arm detects desired object after user keeps his/her finger on top of it. After the lifting phase, the robotic arm tracks the user’s hand to the designated location and sets the object down. The advantages of the hand gesture method are that the object detection process is more accurate and there is no need for gridded working space since the robot has the ability to track user’s finger to any arbitrary position4. On the other hand, the robotic arm is not fully automated as it is required that user interacts with the robot during the whole pick and place operation.     
\subsection{Selected Design}
Choosing a final design is a crucial step for reaching production. However, this can only be achieved once the individual components within the system have undergone a rigorous selection process. During this process components are compared against potentially better alternatives. Each candidate is rated on based on factors that affect the final design. These factors each hold their own weight, where one factor may be more important than another. Within these decision matrices the components are given a score of 0-10 point, where a higher rating is better. 
When deciding on which gripper to use, it is important to take safety into consideration as shown in Table 2. This is the part of the robot that will come into contact with other objects most frequently. Magnets can be seen as very dangerous due to operators potentially having jewelry or metallic surgical implants. The 2-finger gripper is chosen due to its high reliability in picking up and placing an object at a specific point. Choices such as the scooper would essential just be tipping the block on its final location and would not be consistent.
\begin{table}
  \begin{center}
\def~{\hphantom{0}}
  \begin{tabular}{lcccccc}
 & Speed & Accuracy & Precision & Cost & Safety & Rank \etal \\[3pt]

Weighting Factor &	0.125 &	0.125 &	0.125 &	0.350 &	0.275 &	1.000 \\
Suction	& 5 & 6 & 4 & 5 & 7 \\	
Weight	& 0.625	 & 0.75 & 	0.5 &	1.75 & 	1.925 & 	5.550 \\
Scooper	 & 7 &	8 &	4 &	8 &	6 \\	
Weight	& 0.875 &	1 &	0.5 &	2.8 &	1.65 &	6.825 \\
Magnetic	& 10 & 	5 &	7 &	6 &	5 \\	
Weight	& 1.25	& 0.625	& 0.875	& 2.1 &	1.375 &	6.225 \\
2-Finger &	6 &	9 &	9 &	8 &	6 \\	
Weight	& 0.75 &	1.125 &	1.125 &	2.8 &	1.65 &	7.450 \\
Adhesive	& 8 &	7 &	6 &	4 &	9 \\	
Weight	& 1	& 0.875 &	0.75 &	1.4 &	2.475 &	6.500 \\
  \end{tabular}
  \caption{Decision Matrix of Gripper.}
  \label{tab:kd}
  \end{center}
\end{table}
After researching Arduino boards, Table 3 shows that nearly any micro controller boards have nearly the same functionality for the scope of this project, since a servo driver will be used.  This is seen in the decision matrix where although some board have smaller form factors, they have about the same number of analog and digital pins3. Given that the motors will be driven on external power sources and drivers, the only pins require are for digital output to the drivers. Cost is the main deciding factor here where Uno wins due to its availability online third-party retail price around 6.00 USD.
\begin{table}
  \begin{center}
\def~{\hphantom{0}}
  \begin{tabular}{lcccccc}
	& Feasibility	& Analog In/Out	& Digital IO/PWM &	Memory &	Cost &	Rank \etal \\[3pt]
Weighting Factor &	0.125 &	0.125 &	0.125 &	0.275 &	0.35 &	1.000 \\
Nano & (328P) &	7 &	6 &	7 &	7 &	6 \\	
Weight & 0.875 &	0.75 &	0.875 &	1.925 &	2.1 &	6.525 \\
Leonardo &	7 &	8 &	7 &	8 &	5.5 \\	
Weight	& 0.875 &	1 &	0.875 &	2.2 &	1.925 &	6.875 \\
Uno	 & 8 &	8 &	7 &	7 &	9 \\	
Weight &	1 & 1 &	0.875 &	1.925 &	3.15 &	7.950 \\
Micro &	9 &	7 &	7 &	8 &	7 \\	
Weight	& 1.125 &	0.875 &	0.875 &	2.2 &	2.45 &	7.525 \\
Mini	& 6 &	7 &	7 &	6 &	8 \\	
Weight	& 0.75 &	0.875 &	0.875 &	1.65 &	2.8 &	6.950\\
  \end{tabular}
  \caption{Decision Matrix of Arduino Boards.}
  \label{tab:kd}
  \end{center}
\end{table}
Stepper motors and servos will be the muscle of the build, so it is better if they are able to apply more torque. However, when taking into consideration that some motors will have to support the weight of other motors, the weight of the motors itself becomes a huge factor. The compact and lightweight build of the small reduction stepper motor and MG996R pulls them ahead of the other options in Tables 4-5.

\begin{table}
  \begin{center}
\def~{\hphantom{0}}
  \begin{tabular}{lccccccc}
	& Mass & Holding Torque &	Steps per Revolution &	Easy to Set up	& Current Draw & Cost &	Rank \etal \\[3pt]
Weighting Factor &	0.225	& 0.125 &	0.125 &	0.100 &	0.200 &	0.225 &	1.000 \\
NEMA 14-size Hybrid bipolar &	7 &	3 &	5 &	7.000 &	7.000 &	7 \\	
Weight &	1.575 &	0.375 &	0.625	& 0.700 &	1.400 &	1.575 &	6.250 \\
NEMA 17-size Hybrid (bipolar/unipolar) &	5 &	4 &	5 &	4.000 &	4.000 &	6 \\	
Weight	& 1.125 &	0.5 &	0.625 &	0.400 &	0.800 &	1.35 &	4.800 \\
NEMA 23-size Hybrid bipolar &	1 &	5 &	5 &	4.000 &	5.000 &	5 \\	
Weight &	0.225 &	0.625 &	0.625 & 0.400 &	1.000 &	1.125 &	4.000 \\
NEMA 23-size Hybrid (bipolar/unipolar) &	4 &	9 &	5 &	7.000 &	2.000 &	1 \\	
Weight	& 0.9	& 1.125 &	0.625 &	0.700 &	0.400 &	0.225 &	3.975 \\
Small Reduction Stepper Motor &	9 &	1 &	9 &	7.000 &	8.000 &	9 \\	
Weight &	2.025 &	0.125 &	1.125 &	0.700 &	1.600 &	2.025 &	7.600 \\

  \end{tabular}
  \caption{Decision Matrix of Steppers.}
  \label{tab:kd}
  \end{center}
\end{table}

\begin{table}
  \begin{center}
\def~{\hphantom{0}}
  \begin{tabular}{lccccccc}
	
	& Mass &	Speed &	Stall Torque &	Operating Voltage &	Angle Range	 & Cost &	Rank \etal \\[3pt]
Weighting Factor &	0.175  & 0.100 &	0.250 &	0.100 &	0.100 &	0.275 &	1.000 \\
MG996R High Torque &	9 &	8.000 &	6.000 &	7.000	& 8.000 &	8.000 \\	
Weight	& 1.575 &	0.800 &	1.500 &	0.700 &	0.800 &	2.200 &	7.575 \\
SG92R Micro	& 6	 & 7.000 &	9.000 &	6.000 &	6.000 &	6.000 \\	
Weight &	1.05 &	0.700 &	2.250 &	0.600 &	0.600 &	1.650 &	6.850 \\
Futaba S3003	& 7 &	5.000 &	7.000 &	7.000 &	9.000 &	7.000\\	
Weight &	1.225 &	0.500 &	1.750 &	0.700 &	0.900 &	1.925 &	7.000 \\

  \end{tabular}
  \caption{Decision Matrix of Servos.}
  \label{tab:kd}
  \end{center}
\end{table}

Deciding the structural component of the build is crucial as there are huge tradeoffs between varied materials. For a robotic arm, the material needs to be as light as possible, but still be able to handle the loads due to its own weight and the objects to be moved. While balsa would be an excellent choice due to its low density, it would most likely fail due to the loading conditions. Table 6 depicts that acrylic is a great compromise between strength, weight, and cost.

\begin{table}
  \begin{center}
\def~{\hphantom{0}}
  \begin{tabular}{lccccccc}
	
	& Material Cost 	& Machining and Labor Cost &	 Density (Kg/m3) &	Flexural Strength (MPa) &	Rank \etal \\[3pt]
Weighting Factor &	0.275 &	0.175 &	0.425	& 0.125 &	1.000 \\
6061 Aluminum	& 4.000 &	2.000 &	1.000 &	6.000 \\	
Weight &	1.100 &	0.350 &	0.425 &	0.750 &	2.625 \\
Carbon Fiber &	2.000 &	4.000 &	4.000 &	9.000 \\	
Weight	& 0.550 &	0.700 &	1.700 &	1.125 &	4.075 \\
Balsa Wood & 	7.000 &	6.000 &	9.000 &	1.000 \\	
Weight	& 1.925 &	1.050 &	3.825 &	0.125 &	6.925 \\
Acetal &	6.000	& 6.000 &	5.000 &	4.000 \\	
Weight	& 1.650 &	1.050 &	2.125 &	0.500 &	5.325 \\
Acrylic	 & 9.000	& 9.000 & 	6.000 &	3.000 \\	
Weight	 & 2.475 &	1.575 &	2.550 &	0.375 &	6.975 \\

  \end{tabular}
  \caption{Decision Matrix of Materials.}
  \label{tab:kd}
  \end{center}
\end{table}

The performance accuracies of the infrared and ultrasonic sensors were tested using varied materials in order to come to a concrete decision. In this experiment, the two sensors were attached to a micro controlled RC car moving at constant speed. The sensors measured the distance between the RC car and a still object over time. The following graphs show the accuracy of detection between the two sensors amongst wood, plastic, and rubber.

\begin{figure}
  \centerline{\includegraphics{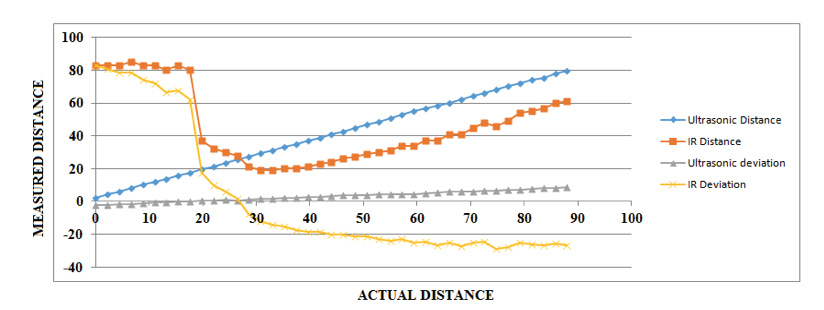}}% Images in 100% size
  \caption{Actual Distance (cm) vs. Measured Distance (cm) for Wooden Object.}
\label{fig2}
\end{figure}

\begin{figure}
  \centerline{\includegraphics{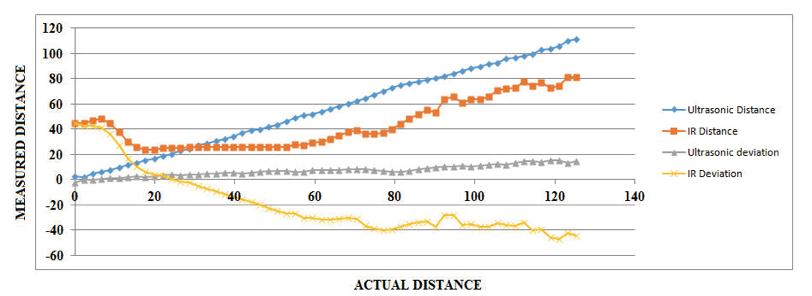}}% Images in 100% size
  \caption{Actual Distance (cm) vs. Measured Distance (cm) for Metallic Object}
\label{fig3}
\end{figure}

\begin{figure}
  \centerline{\includegraphics{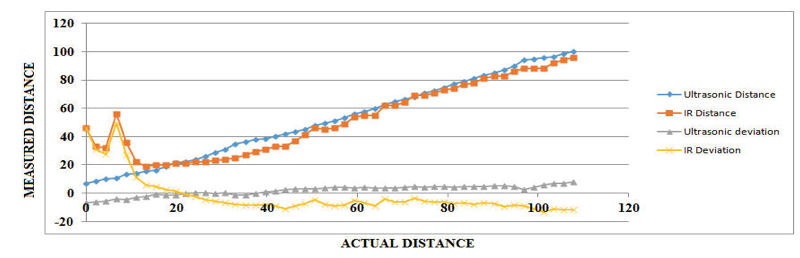}}% Images in 100% size
  \caption{Actual Distance (cm) vs. Measured Distance (cm) for Rubber Object.}
\label{fig4}
\end{figure}

It became clear in Table 7 of choosing a vision system that if there was enough money to allocate toward the better. The vision system are the robot’s eyes to the world, using an inferior option due to price could result in very complex coding or just missing information when attempting to locate items. The PixyCam is clearly the best choice due to its ease of use and onboard processor for quickly identifying objects. Additionally, it has its own Arduino library to simplify coding.
\begin{table}
  \begin{center}
\def~{\hphantom{0}}
  \begin{tabular}{lccccccc}
	
	&Speed	& Accuracy &	Interference &	Cost &	Safety &	& Rank \etal \\[3pt]
Weighting Factor &	0.15	& 0.15	& 0.2 &	0.300 &	0.2 &	1.000 \\
Ultrasonic (HC-SR04) &	9 &	7 &	5 &	6 &	9 \\	
Weight &	1.35 &	1.05 &	1.0 &	1.8 &	1.8 &	7.000 \\
Infrared(SHARP GP2Y0A21YKOF) &	9 &	9 &	6 &	5 &	7 \\	
Weight &	1.35 &	1.35 &	1.2 &	1.5 &	1.4 &	6.800 \\
Webcam(Logitech c270) &	6	& 8 &	8 &	2 &	7 \\	
Weight &	0.9 &	1.2 &	1.6 &	.6 &	1.4 &	5.7 \\
PixyCam	& 10 &	10 &	10	& 1 &	9 &	7.1 \\
Weight &	1.5 &	1.5 &	2.0 &	.3 &	1.8 &	7.1 \\

  \end{tabular}
  \caption{Decision Matrix of Type of Sensor.}
  \label{tab:kd}
  \end{center}
\end{table}

\section{ Methodology}
Though a prototype has not been made and tested, a final design has been selected. Our design at the moment includes the layout of the arm and some preliminary coding. Since we do not have a finished prototype yet, no testing has been done which would have included stress testing in ANSYS. However, the decision process alone involved many technical considerations. The process for choosing the end design involved many methods; these methods were mainly used to determine the best option for the project. In this section those methods are outlined, including our progress towards a finished prototype. 
\subsection{Calculations}
Though a prototype hasn’t been made and tested, a final design has been selected. Our design at the moment includes the layout of the arm and some preliminary coding. Since we do not have a finished prototype yet, no testing has been done which would have included stress testing in ANSYS. However, the decision process alone involved many technical considerations. The process for choosing the end design involved many methods; these methods were mainly used to determine the best option for the project. In this section those methods are outlined, including our progress towards a finished prototype. 
\begin{figure}
  \centerline{\includegraphics{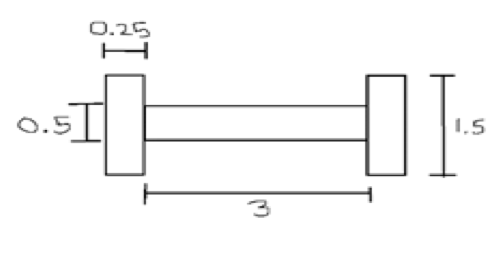}}% Images in 100% size
  \caption{I-beam Cross Section}
\label{fig5}
\end{figure}

Though a prototype hasn’t been made and tested, a final design has been selected. Our design at the moment includes the layout of the arm and some preliminary coding. Since we do not have a finished prototype yet, no testing has been done which would have included stress testing in ANSYS. However, the decision process alone involved many technical considerations. The process for choosing the end design involved many methods; these methods were mainly used to determine the best option for the project. In this section those methods are outlined, including our progress towards a finished prototype. 
When making the design for the arm an analysis of the stress had to be done. More specifically, the stress that each of the links of the arm would undergo. Doing so gives us an idea of what material we can use as well as how the weight of the motors affects the design. Some assumptions had to be made to simplify this calculation: the links have an I-shaped cross section with these dimensions (Fig 1) and the links are cantilever beams with a mass on the free end. To find the moment you use equation 1; the moment is 0.49Nm:
\begin{equation}
  M = mgd
  \label{Helm}
\end{equation}
To find the max bending stress you use equation 2 and 3; the max bending stress is 2.2MPa:
\begin{equation}
  I=(b_1 h_1^3 )+(b_2 h_2^3 )+(b_3 h_3^3)
  \label{Helm}
\end{equation}
\begin{equation}
  \sigma=My/I
  \label{Helm}
\end{equation}

The total stress would normally include bending, normal, torsion, and transvers. However, since the bending stress is significantly larger than the other ones, it can be assumed that the total stress is equal to the max bending stress. Once the requirements are fully known for the range of the arm, the actual dimensions of this arm can be drawn up. Using the equations provided, the stress again can be calculated.        	
Originally the degree of freedom (DOF) for this project was set at 6. This is no longer a design restraint. Designs for arm were made in a way that the arm could complete its task but,  have the smallest amount of DOF. Each DOF requires one motor. Motors are the most expensive part of this design other than the board. Since there is a budget limit, using less motors leaves us with more money to spend on other parts of the design. Calculating the DOF is done using equation 4:
\begin{equation}
  M=3(L-1)-2J_1-J_2
  \label{Helm}
\end{equation}
The proposed design has 5 links (L), 4 full joints ($J_1$), and 0 half joints ($J_2$) resulting in 4 DOF. With the motor for gripper included, this arm design requires 5 motors. As can be seen, reducing the number of links and joints simplifies the design. 
The gripper that we created uses gears. When assembling the gripper, the gears had to be spaced appropriately so that they would work smoothly together. The distance between the centers of both gears can be calculated using equation 5 and 6: 
\begin{equation}
 Pitch Diameter=  (Number of teeth*Circular Pitch)/(Number of teeth+2)
  \label{Helm}
\end{equation}
\begin{equation}
Center Distance=  (Pitch Diameter_{Gear1}+Pitch Diameter_{Gear2})/2
  \label{Helm}
\end{equation}
These equations were used for just a mock design of the gripper, but can be used again once actual dimensions are set. 
Though the gear gripper seems to be the best option, a different simpler design is proposed. This design involves using a servo motor to pull back the plunger of a syringe that is connected to a suction cup via plastic tubing. Assuming the suction cup is in complete contact with an object, a vacuum force is created. This design is simple and considerably cheaper than the gear gripper. However, the strength of this gripper is uncertain. What is the maximum force of this vacuum and what weight can it thus support?
It is assumed that temperature of the air is constant throughout the process, the syringe has a volume of 60mL with a 20mm diameter, the suction cup is a hemispherical shape with a diameter of 30mm, and that the plunger of the syringe is retracted 4.76cm. The volume of the suction cup is found by using equation 7:
\begin{equation}
V_1=2/3 \pi r^3
  \label{Helm}
\end{equation}
Using equation 7 the volume of the suction cup is found to be $7.06 \times 10^{-5} m^3$. This is the initial volume of the system (the plunger is not pulled back yet). The volume of the plastic tubing is ignored because it is so small. The final volume of the system is described in equation 8:

\begin{equation}
V_f=V_1+V_2
  \label{Helm}
\end{equation}
$V_2$ is the volume of the syringe with the plunger pulled back. The syringe is a cylinder and its volume is defined by equation 9:
\begin{equation}
V_2=\pi r^2 h
  \label{Helm}
\end{equation}
Using the radius of the syringe and the length of retraction for h, $V_2$ is $5.98 \times 10^{-5}m^3$. By using equation 8 the final volume equals to $2.2 \times 10^{-5}m^3$. When the plunger is retracted, the pressure changes as a result of volume increasing. To find the final pressure equation 10 is used:
\begin{equation}
P_1 V_1=P_2 V_2
  \label{Helm}
\end{equation}
Assuming that $P_1$ is atmospheric pressure ($1.035x10^5 Pa$), P2 is calculated to be 32,524.13 Pa. With pressure, the force can be found. Equation 11 defines this relation:
\begin{equation}
P=F/A
  \label{Helm}
\end{equation}
Using $P_2$ and $7.07x10^{-4}m^3$ for A, the resulting force is equal to 22.98N. The approximate weight of this force reading is calculated using equation 12: 
\begin{equation}
W=F/9.81
  \label{Helm}
\end{equation}
From equation 12 the weight is 2343.31kg. The approximate weight of a Jenga block is 0.011kg. Therefore, this design in theory should be able to pick up the Jenga block easily. This calculation does leave out the external forces caused when the device will be moving around. It should be noted that if at any time the vacuum is broken (space between suction cup and Jenga block is created) the whole system will fail. This space can be cause by moving the device around rapidly. 
\subsection{Technical Drawings}

These equations were used for just a mock design of the gripper but can be used again once actual dimensions are set. 
\begin{figure}
  \centerline{\includegraphics{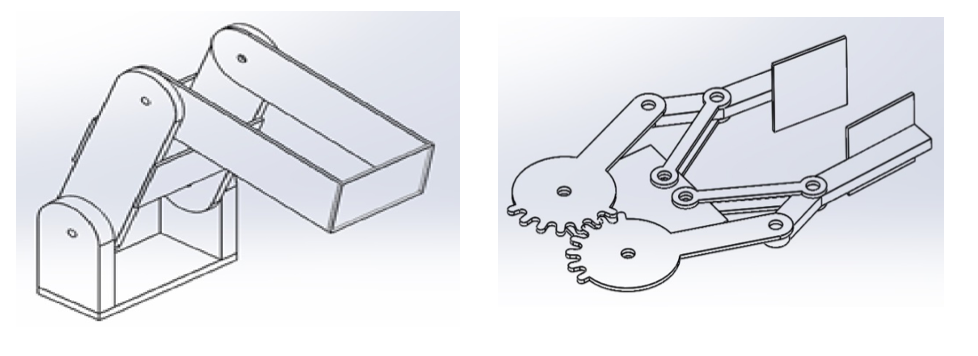}}% Images in 100% size
  \caption{SolidWorks Design of a) Three-Link Arm b) 2-Finger Gripper }
\label{fig3}
\end{figure}

\begin{figure}
  \centerline{\includegraphics{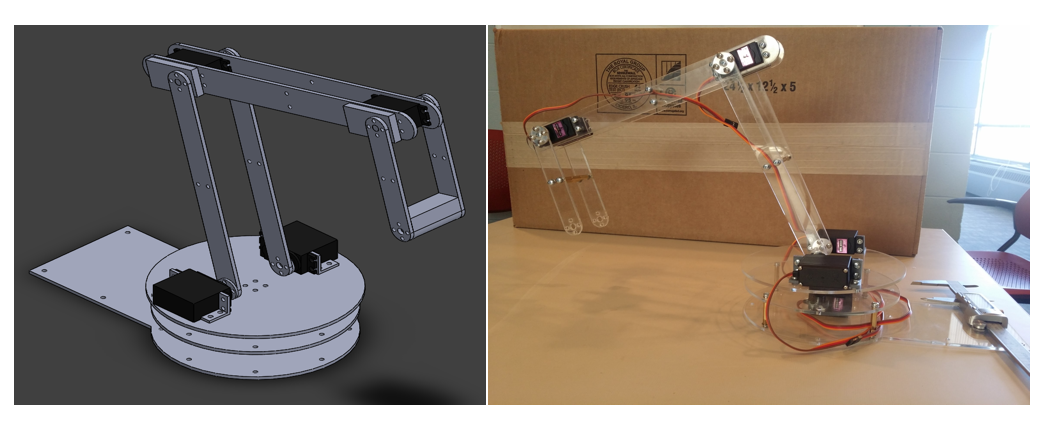}}% Images in 100% size
  \caption{ c) Current Design w/o Gripper and Screws. d) Current Assembly }
\label{fig3}
\end{figure}
Figure 6a shows the arm design without the wrist and gripper attached. The wrist would add another link and joint. With these, the DOF of the arm is 4 like was shown previously in in equation 4. In Figure 6b the gripper design is shown. Notice that the gears don’t have teeth going all around it. This is done so that the gripper’s range when it opens is limited. This ensures that it doesn’t over rotate and malfunction. 
\subsection{Algorithms}
\begin{figure}
  \centerline{\includegraphics{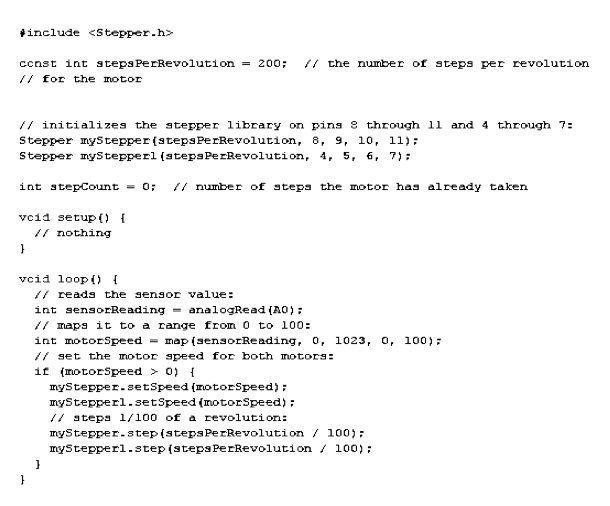}}% Images in 100% size
  \caption{Code for Running Two Stepper Motors. }
\label{fig3}
\end{figure}
Our final design will consist of multiple motors that will all have to work together to move the arm. This coordination is all done by the code written for the Arduino. Using multiple motors together is complex, so through this semester we’ve been building up to it. The arm will primarily use stepper motors. Only the gripper uses a servo motor. Two stepper motors being controlled by a single board can be seen in Figure 4. In this code, a potentiometer is used to control the rotational speed of both motors. The motors rotate clockwise, a reading is received from the sensor, the reading is mapped on a range from 0 to 100, and then the motors rotate at that speed. This code taught us how to control the speed of the stepper motors. 
\subsection{Algorithms}
\begin{figure}
  \centerline{\includegraphics{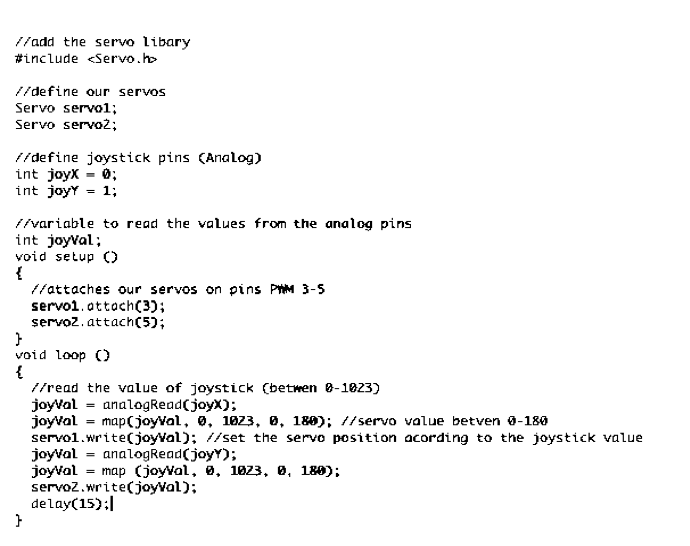}}% Images in 100% size
  \caption{Code for Controlling Servo by Joystick.}
\label{fig3}
\end{figure}
The first step is to be able to manually control multiple motors and the final goal is to have them move somewhat autonomously. A code was used to control two servos using a joystick (Figure 5). Moving the joy stick left or right will move one servo clockwise and counterclockwise. Moving the joy stick up or down will move the other servo clockwise or counterclockwise. Our first prototype will be controlled in a similar way until we figure out the code to make it more autonomous. 
\subsection{Algorithms}
\begin{figure}
  \centerline{\includegraphics{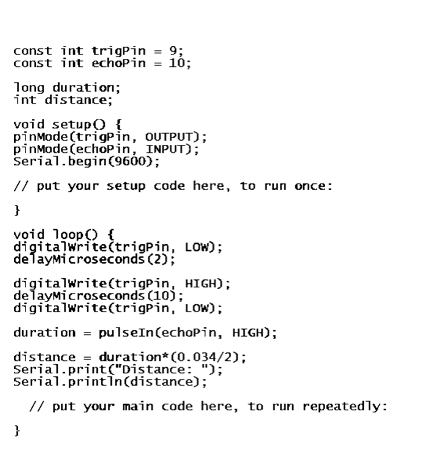}}% Images in 100% size
  \caption{Code for Ultrasonic Sensor.}
\label{fig3}
\end{figure}
One of the components of making this pick and place more autonomous is adding sensors. The ultrasonic sensor emits an ultrasound that travels through the air and if it contacts something will bounce back to the sensor. By using the speed of the ultrasound and the time it took to travel, a distance can be calculated. The code shown in Figure 6 controls an ultrasonic sensor and outputs the distance. In our final design the ultrasonic sensor can work in combination with an infrared sensor that can detect the shapes of objects. 
\subsection{House of Quality}
The quality function deployment (QFD) is made by crossing customer requirements and functional requirements. By doing so the most important criteria can be found. For this project, energy requirement and motor type were found to be the most important (had the most relevance in every aspect of the project). Energy requirement in our case mainly refers to how much power the motors require and what Arduino boards can handle the amount of current we will be using. Motor type here refers to which type of motor to use for a certain function (servo or stepper). As you can see, the motors are basically the most vital component of this project. Most future work is going towards motors.

\begin{figure}
  \centerline{\includegraphics{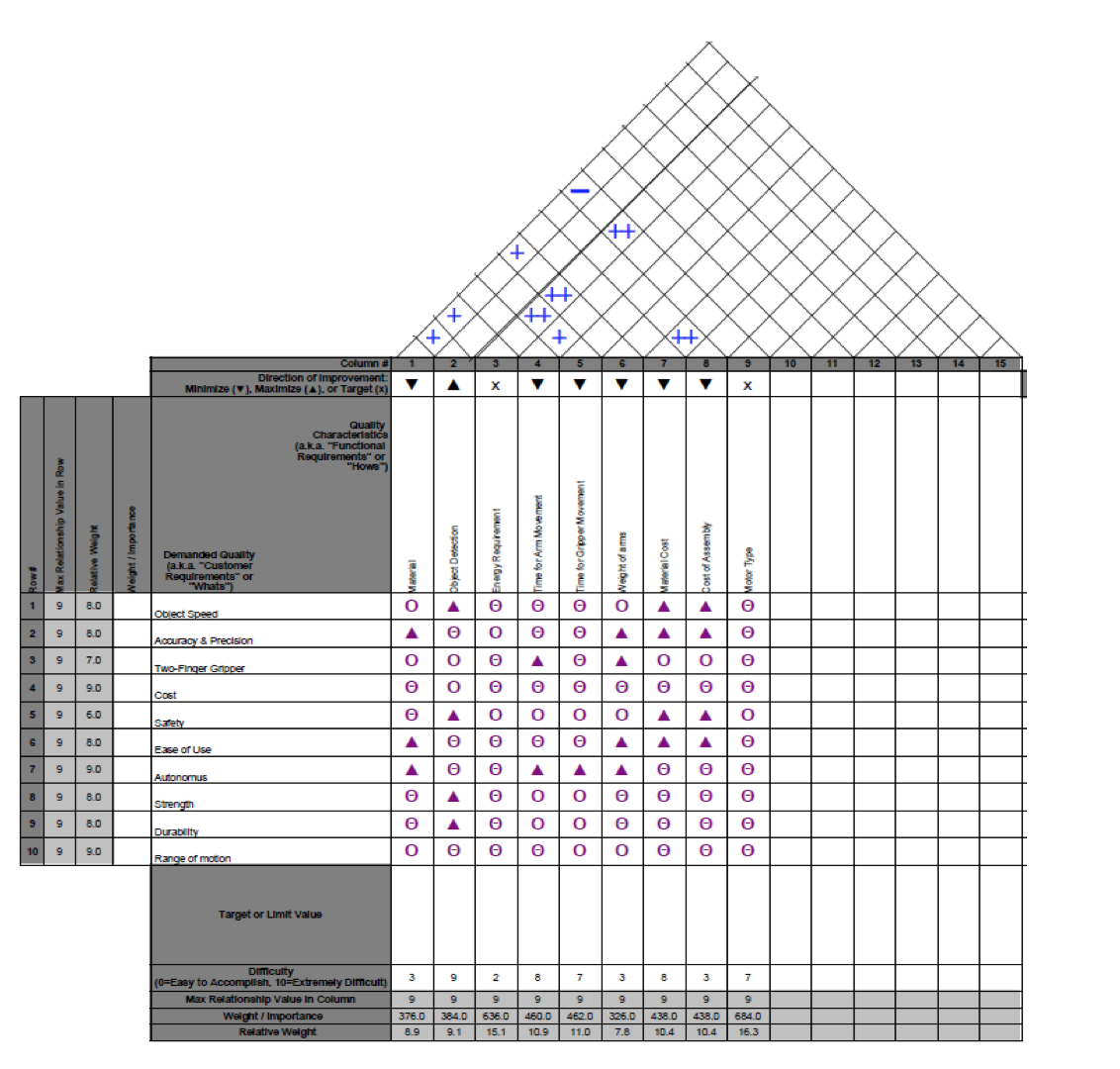}}% Images in 100% size
  \caption{House of Quality for Pick and Place Robot Arm.}
\label{fig3}
\end{figure}

\subsection{Method}
The first step in designing the arm, is understanding the scale for which the arm will be operating in. Given that this project was intended to test the designers’ ingenuity, a lot of areas are left up to interpretation to give room for the best final product to be created. The first assumption made is that since the product is an “arm”, one should make it roughly the same size of a human arm. The initial idea was to make the arm the length of the average human 30+ inches7. This length would prove to not be achievable given the weight of the arm, the objects, and the budget for buying the motors to support all the weight. 
Given that the motors are the muscles behind the build, it only made sense to focus the build around the motors. The initial though was to use stepper motors as they have a higher holding torque. However, their torque drop significantly while in motion, which is not desirable in a dynamic arm. Also, upon sourcing stepper motors, they were generally 10 times the mass of a servo of equivalent maximum holding torque. Ultimately, the servo was the final decision. The servo chosen in the MG996R a revamp version of the commonly used MG995 that allows for better positional accuracy. It features a high stall torque of 9.4 kg*cm at 4.8V, in compact package at only 55g. The reason for its popularity is it cost to performance ratio. Other motors with equal or higher max torques are pale in comparison when it comes to value.
Under the assumption of the MG996R, the arm length was revisited. Static calculations were used to test the maximum torque each motor would undergo. Each motor would be subject to the weight of any successive linkages, motors, grippers, or object. Once it was confirmed that the motor would not fail under static loading, a factor of 1.5 was implemented into the applied torque to account for dynamic loading (i.e. moments of inertia). Varying the lengths of the arm concluded that a maximum length for the arm at full horizontal extension should be 24 in. With this length, the most likely point of failure would be the base motor. However, two motors can be used on the base to help this. Additionally, 24in is the maximum reach of the arm, and not the operating conditions, so the arm will undergo significantly less stress during actual operations. These preliminary calculations are solely for defining limitations.
Given the limitations of the arm, the operating conditions can be specified for safe operation. In order to no put too much stress on the arm, but still utilizing the power of them, it will be assumed that the arm will not have to reach further than 24” at 45° to the base. This equates to the end of the arm to never need to go further than 16.97” horizontally from the base. It is assumed the arm will be mounted to the edge of a desk, so it will only have the area in front of it to work with. The effective workspace for the arm can be defined by the half cylinder shown in Figure 11, where the radius (r) is equal to the length (L) at 16.97” and the arm base is centralized at the origin. The arm should be able to pick n’ place anywhere within this workspace, besides near the origin do to interferences with the base.
\begin{figure}
  \centerline{\includegraphics{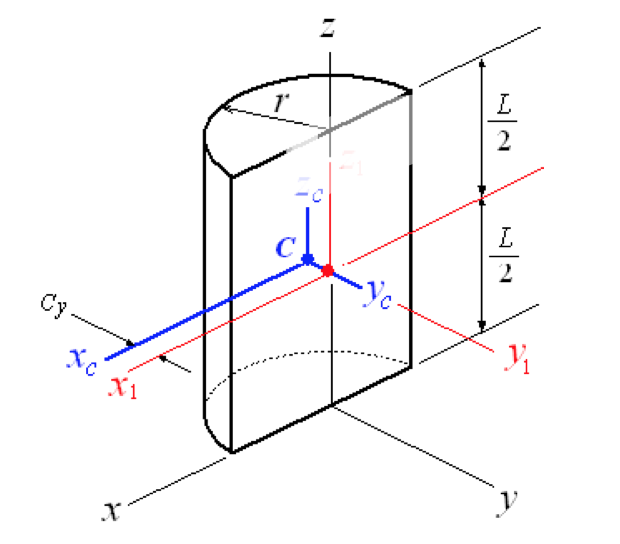}}% Images in 100% size
  \caption{Work Space for Pick and Place Robot Arm}
\label{fig3}
\end{figure}

The methods employed revolved around constant process of observation, ingenuity, implementation, and reiterating the design. Using the HoQ as seen in Figure 10 is used to define a set of parameters to break down the project into conquerable sections. The HoQ parameterized ten essential customer requirements that need to be taken into consideration when looking for the means of completing the design.
The speed of the entire operation is based upon two limiting factors, the speed of the motors and processing speeds of the microcontroller and vision system. While running the motors at max speed may seem like a clever idea, it is actually bad idea, because as their speed increased their holding torque will decrease. The processing speed of the Arduino Uno and PixyCam are more than sufficient for this application, as the initial and final positions of object placement are stationary. Therefore, a high polling rate is not necessary. The major limiting factor to the speed of the arm is optimizing the tradeoff between speed and holding torque of the MG996R servo motors.
Achieving maximum accuracy and precision positioning during operation is the most complex element. Accuracy/Precision are functions of factors such as vibration, momentum and inertias, and signal noise. Proportional-Integral-Derivative (PID) controls will be implemented as the last step to fine tune the arm for smoother and more predictable motion. 
Picking up and placing the object will be accomplished by means of a pinching “two-finger “gripper. The two-finger gripper allows for more complex geometries to be picked up. Whereas, a vacuum/suction style gripper would be limited to flat surfaces with a minimum area equal to that of the suction cup. The gripper also allows for easy programing of the force required to stop the object from slipping out the gripper once the object is picked up. The normal force can be increased by sending more power to the servo motor driving the gripper. This allows the arm to adapt to a wide range of surface frictions for each object. 
Cost is a limiting factor of all projects and must always be in mind when determining the design and implementation of a product. Specifically, this project has a 250 USD budget for parts. Currently the cost of all the parts is 199.25 USD.
Safety risks are very minimal in this design. The arm doesn’t have a large mass, dimensions, nor is it moving at high speeds. The operating voltage is about 5V with a maximum currant draw of about 12A, so electrocution is not a problem. The main point of concern will be the motors overheating/stalling and ultimately causing the arm to fall. As for human interaction safety precautions, such as the arm swinging into an individual, an emergency stop button will force the arm to immediate stop all movement and hold its current position until told further actions.  
\section{Preliminary Conclusions / Future Work}

There are several things that need to be completed still. All hardware tasks have been completed. The CAD models have been drawn and manufactured. Any sourced parts have been purchased, arrived, or are in transit. A physical point of concern is the structural integrity of the base motor. There is minor tilting due to the weight of the arm structure. A fix will be implemented by installing a plastic cylinder with a layer of Teflon between the plastic and rotating acrylic plate. This will prevent the tilting while minimize energy losses due to friction. The rotation pegs need to be reworked and printed. The next step in the process is to finalize the coding in Arduino. Once a PixyCam arrives, the vison dependent kinematic equations can be formatted into Arduino code. Also, after the arm is moving properly, the last step is integrating PID controls for fine tuning. In preparation for Expo, a workspace and necessary items such as the grippers need to be sourced. This work must be done in the next two weeks to realistically have a successful prototype.

\bibliographystyle{jfm}
\bibliography{jfm-instructions}

\end{document}